\documentclass{article}
\usepackage{spconf,amsmath,graphicx}
\usepackage{booktabs}

\title{END-TO-END LIP SYNCHRONISATION BASED ON PATTERN CLASSIFICATION}

\name{You Jin Kim$^1$, Hee Soo Heo$^1$, Soo-Whan Chung$^{1,2}$ and Bong-Jin Lee$^1$}
\address{
  $^1$Clova AI, Naver Corporation, Seongnam-si, Gyeonggi-do, South Korea\\
  $^2$Department of Electrical \& Electronic Engineering, Yonsei University, Seoul, South Korea}

\begin{document}
%
\maketitle
\begin{abstract}
The goal of this work is to synchronise audio and video of a talking face using deep neural network models.
Existing works have trained networks on proxy tasks such as cross-modal similarity learning, and then computed similarities between audio and video frames using a sliding window approach. 
While these methods demonstrate satisfactory performance, the networks are not trained directly on the task.
To this end, we propose an end-to-end trained network that can directly predict the offset between an audio stream and the corresponding video stream. The similarity matrix between the two modalities is first computed from the features, then the inference of the offset can be considered to be a pattern recognition problem where the matrix is considered equivalent to an image. The feature extractor and the classifier are trained jointly.
We demonstrate that the proposed approach outperforms the previous work by a large margin on LRS2 and LRS3 datasets.

\end{abstract}
\section{Introduction}

Audio-to-video synchronisation is one of the crucial characteristics that any video clip should have. An off-sync video is not only unnatural to the human eye~\cite{bt1359}, but also causes errors in audio-visual speech enhancement~\cite{Ephrat18,Afouras18} and in making datasets for speaker recognition~\cite{Nagrani17} and lip reading~\cite{Chung16}.
However, in several stages of recording and broadcasting, a video can go out of synchronisation. 
During recording, the delay between the audio and the visual stream can cause the video to go out of sync. 
Also, when a video stream is transmitted, loss or corruption of frames can occur during both wired and wireless communication, causing synchronisation errors.      

Cross-modal learning has received increasing attention recently, using two or more data modalities such as text, image, audio and video~\cite{korbar2018cooperative,Owens18, kim2018learning,morishima2002audio}.
It effectively represents not only an instance of a modality, but also the complex correlation between instances of multiple modalities, projecting them into a joint embedding space~\cite{Nagrani18b,Nagrani20d,nagrani2018seeing}.
There are a number of works that have utilised the joint embedding vectors to tackle various tasks.
Speech enhancement~\cite{Ephrat18,Afouras18,afouras2019my}, source localization~\cite{arandjelovic2018objects,senocak2018learning}, active speaker detection~\cite{chakravarty2016cross,chakravarty2016active,garg2000audio}, lip reading~\cite{Assael16,Afouras19,afouras2019asr} and action recognition~\cite{arabaci2018multi,gao2019listen} are examples of audio-video tasks; and
video captioning~\cite{wu2017deep,jin2016video} and video-text retrieval~\cite{sivic2003video,mithun2018learning} are video-text modality tasks.

In this paper, we propose a method based on the joint embedding space to tackle the AV synchronisation problem. More specifically, we predict the time offset between audio and visual streams by using the similarities between the embeddings of the two streams. 


A previous work of particular relevance to this paper is~\cite{Chung16a}, which introduced a two stream architecture called {\em SyncNet} to handle the AV sync task. 
SyncNet is trained to maximise the similarities between features of the audio and the video segments that come from the same point in time, while minimizing similarities for segments that come from different points in time.
In~\cite{chung2019perfect, chung2020perfect}, cross-modal embeddings are learnt by using the multi-way matching objective that optimises the relative similarities between multiple audio features and one visual feature. 
Both use the trained network to project audio and visual inputs into the joint embedding space, and use the sliding window approach to correct the AV sync error. 
By sliding the window of fixed length, the similarities of every embedding pair are calculated, and the sync offset is determined from the embedding indices where the similarity is maximised. 
\cite{halperin2019dynamic} addresses the problem of synchronising lip motion in re-dubbed or animated videos, where dynamic time warping is needed to compute the offset at frame-level. While this is an interesting problem in its own right, we do not consider this case.

In this paper, we propose a new approach to determine AV time offset, by focusing on the fact that the offset is consistent in a {\em recorded} video. 
We assume that this consistency can be represented as a linear pattern in the similarity matrix, calculated from continuous audio and visual features. 
The matrix consists of similarities between all audio-visual embedding pairs extracted from the two-stream convolutional neural network (CNN) feature extractor. 
Based on this assumption, we formulate AV synchronisation task as a pattern classification problem where the offset value can be predicted directly. 
Pattern matching approach integrates the information over multiple time steps, learning to ignore the non-discriminative time steps such as when mouth is occluded or phonemes are repeated.
In addition, this method enables end-to-end training with feature extractors. 
We demonstrate the proposed classification-based approach outperforms the existing method in lip synchronisation.


\begin{figure}[t]
  \centering
  \includegraphics[width=0.85\linewidth]{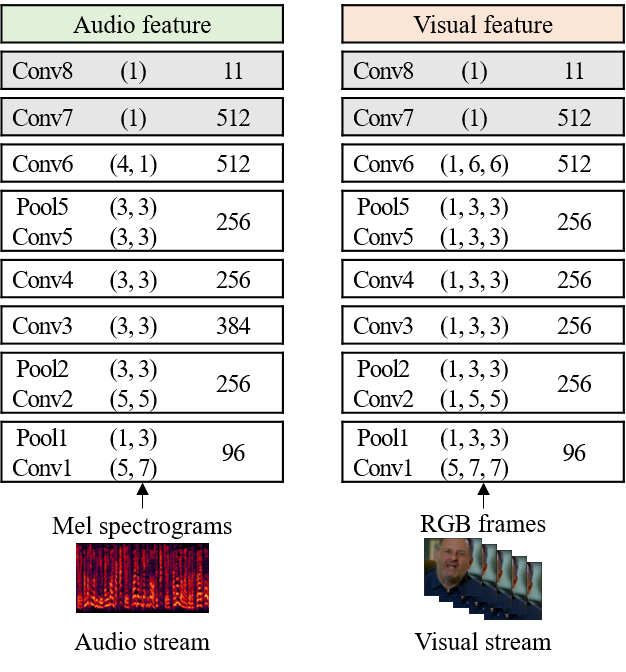}
  \caption{Architecture of feature extractor. Layer, kernel size (T, H, W), and the number of filters are denoted.}
  \label{fig:3}
\end{figure}


\section{Cross-modal learning framework}

In this section, we introduce the baseline system and propose a novel framework for directly predicting the temporal offset between audio and visual streams.
Typically, a process of AV synchronisation is as follows: (1) a network configuration is defined to handle audio and visual streams (2) the network is trained to learn audio-visual representations in a joint embedding space (3) audio-visual offset is established using the output of the trained network. Each stage of the pipeline is described in the following subsections.

\subsection{Network architecture}

The network specifications are similar to~\cite{chung2019perfect} except for the small change in the audio stream. Figure~\ref{fig:3} shows the layer configuration for both audio and visual streams. The inputs to the network are audio and visual streams of a synchronised talking face video. 0.2 seconds of the clip are digested to the each stream of the feature extractor at each timestep.

\subsubsection{Audio stream}

The input to the audio stream is 40-dimensional mel-spectro\-gram extracted from the 16kHz audio signal. 
As a result, the input dimension is $40\times20$ where the 20 frames come from the 0.2-second input length of the audio. 
The architecture of the audio stream is based on VGG-M~\cite{Chatfield14} that is originally designed for the image classification, but the filter sizes are modified to accommodate the smaller input dimensions of mel-spectrogram. 

\subsubsection{Visual stream}

For the visual stream, the input is the RGB frames extracted from cropped talking face videos. 
The frames are extracted from the video at 25 fps, and then resized into $224\times224$. 
The network ingests 5 frames at a time (corresponding to 0.2 seconds at 25 fps). The dimension of the input is therefore $224\times224\times3\times5$ ($H\times W\times C\times T$). As with the audio stream, the architecture is based on the VGG-M~\cite{Chatfield14} network, except that a 3D filter instead of the 2D one at the first layer to capture the temporal information through the stacked frames.

\begin{figure}[ht]
  \centering
  \includegraphics[width=0.65\linewidth]{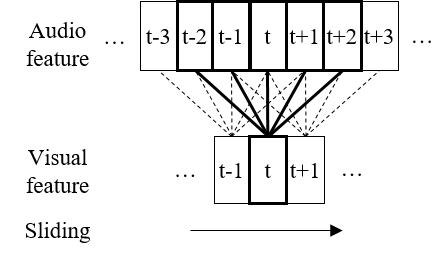}
\caption{Example of sliding window approach. In this figure, the size of a window is set to $5$ for simplicity.}
\label{fig:2}
\end{figure}

\begin{figure*}[ht!]
\begin{center}
    \centering
    \includegraphics[width=1.0\linewidth, trim = 0 5 0 0, clip]{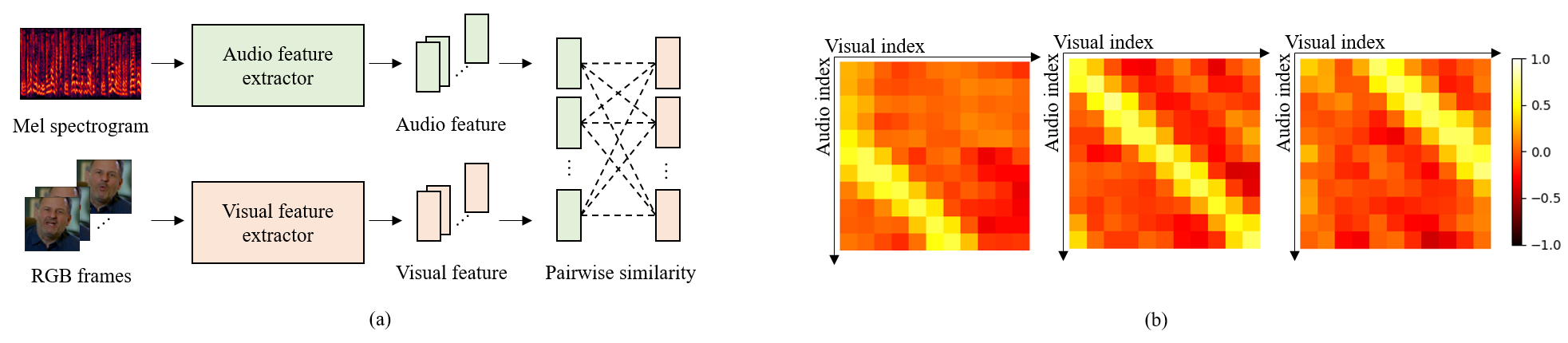}
\caption{(a) Process of similarity matrix generation. The matrix consists of pairwise similarities between audio and visual features. (b) Examples of the similarity matrix. Audio and visual streams are synchronised (middle), the visual stream leads the audio (left), and the opposite case (right).}
\label{fig:ss_application}
\end{center}
\end{figure*}


\subsection{Learning cross-modal embeddings}

The goal here is to train audio and visual representations in a joint embedding space, such that matching audio-visual contents lies close together in the embedding space, while non-matching contents lie far apart.

To achieve this goal, SyncNet uses the contrastive loss from Siamese networks~\cite{chopra2005learning} to train the feature extractor. The loss function compares one audio and one visual features, and calculates Euclidean distance between them. 
It minimises the distance (maximises the similarity) between audio and visual features that are matched, and
maximises the distance (minimises the similarity) for a non-matching features.
The loss is defined as follows: 
\begin{equation}
  L = \frac{1}{2N}\sum_{i=1}^{N}(y_{i})d_{i}^{2} + (1 - y_{i})max(margin - d_{i}, 0)^{2},
  \label{eq:1}
\end{equation}
\noindent where $d_{i}$ is the Euclidean distance between the $i$-th audio feature and $i$-th visual feature, $y_{i}$ is true label that has $1$ for a matching pair and $0$ for a non-matching pair, and $N$ is mini-batch size.

\cite{chung2019perfect, chung2020perfect} have proposed a multi-way matching objective that enforces the {\em relative} similarity of a matching pair over multiple non-matching pairs. 
The similarity between audio and visual features are represented by the inverse of Euclidean distance, and passed to softmax layer. The network is trained using the relative similarity and cross-entropy loss.
Specifically, the loss function is defined as follows:
\begin{equation}
  L = - \frac{1}{N}\sum_{i=1}^{N}log\frac{exp({d}_{i, i})}{\sum_{j=1}^{A}exp(d_{i, j})},
  \label{eq:2}
\end{equation}
\noindent where \(d_{i, j}\) is the inverse of Euclidean distance between the $i$-th audio feature and the $j$-th visual feature, and $A$ denotes the number of the audio features. 
The method assumes that the audio and the visual streams are synchronised in the training data, which can be assumed to be the case in LRS2 and LRS3 datasets.

We use an angular variant of the cross-modal matching loss~\cite{chung2020seeing} which has shown to generalise better to new domains, compared to networks trained with the Euclidean metric. 
The loss is defined as follows:
\begin{equation}
  L = - \frac{1}{N}\sum_{i=1}^{N}log\frac{exp(w \cdot {s}_{i, i}+b)}{\sum_{j=1}^{A}exp(w\cdot s_{i, j}+b)},
  \label{eq:3}
\end{equation}

\noindent where \(s_{i, j}\) is cosine similarity between the $i$-th audio feature and the $j$-th visual feature that are normalized. $w$ and $b$ are learnable parameters to adjust the dynamic range of the distances since the cosine function is bound to $[-1,+1]$.


\subsection{Computing the audio-visual offset}

\subsubsection{Baseline}
To determine the time offset from AV stream, \cite{Chung16a} and \cite{chung2019perfect} use the sliding window approach shown in Figure~\ref{fig:2}. 
In a fixed-length window, the distances between a visual feature and all audio features contained in a window are calculated. 
The audio representations are zero-padded outside of the length of the input segment, which can adversely affect performance.
It is not sufficient to determine the time offset using a single window due to the limited phonetic information in the short time period.
For example, there can be a moment when the speaker covers his/her mouth or stops talking, in which case the video cannot provide discriminative information, considering that AV synchronisation is computed based on the phonetic information. 
Therefore, the distances are computed several times by sliding the window and then averaged to obtain more accurate results. The audio and the visual streams are considered synchronised when the feature distances between them are minimised. More details on the sliding window approach can be found in~\cite{chung2019perfect}. 

\subsubsection{Proposed approach}
The proposed method is motivated from the fact that the AV time offset of a video clip is constant within a clip, which can be represented by a linear pattern in a similarity matrix.
We calculate the cosine similarity matrix of size $N\times N$, where $N$ is the number of audio and visual features. The element at $(i,j)|_{i,j<N}$ is the cosine similarity between the $i$-th audio feature and the $j$-th visual feature.

The elements at the point where the audio and the visual features are synchronised have the relatively large value compared to others in the matrix. Since the AV sync error is assumed to be constant in a video clip, the points with the larger similarities should represent a linear pattern in the matrix.
Based on this assumption, we treat the similarity matrix as an image, and use it as an input to a pattern classifier to predict the temporal offset. 
We expect that the classifier extracts the pattern in the matrix, and maps the pattern to a class representing the offset value.

Figure~\ref{fig:ss_application}-(a) shows the process of generating the similarity matrix, and Figure~\ref{fig:ss_application}-(b) shows examples of the patterns for various offset values on similarity matrices. 
If AV are synchronised, the linear pattern can be found along the diagonal of the distance matrix ({\em middle}). If the visual stream leads the audio, the pattern is located parallel to, but below the diagonal ({\em left}), and in the opposite case, it is located above the diagonal ({\em right}). 
Therefore, we hypothesise that it is possible to predict the offset value by formulating the problem as one of pattern recognition.
In our experiments, we search for the offset over $[-5,+5]$ frame range. This makes it a 11-way classification including zero offset. A negative offset denotes the visual stream leads the audio stream, and zero offset represents perfect synchronisation.

We propose two strategies to determine the AV offset based on the pattern recognition approach: (1) the similarity matrices are pre-extracted and the offsets are determined as a classification task, (2) the feature extractor and the offset classifier are trained following end-to-end manner with full supervision. We refer to the former as \texttt{Sync-cls}, and the latter as \texttt{Sync-e2e}.
By training a classifier with a large amount of data, \texttt{Sync-cls} can learn to deal with non-regular patterns that are difficult to detect using the heuristic in the baseline method. 
In addition, \texttt{Sync-e2e} optimises the entire network for further improvement in performance. 
Figure 4 shows an overview of the proposed approach compared to the existing approach.

Table 1 shows the layer configuration of \texttt{Sync-cls} that consists of four convolution layers. 
The $(3 \times 3)$ kernel of the first convolution layer captures the local pattern, while the second layer captures the global pattern in the similarity matrix. 
The second layer's kernel size, $(N \times N)$ is same as the input tensor.
Batch normalisation and ReLU activation are applied respectively after all convolution layers, except for the last layer. 
Note that no stride and max-pooling layer are used, since location invariance would be an adverse characteristic of the network for this task, unlike in image recognition.
Softmax is applied as the last activation function, and cross-entropy loss is used to train the network.
For \texttt{Sync-e2e}, we train the feature extractor and the \texttt{Sync-cls} at the same time, using the same cross-entropy loss.

In addition to the above, we provide another ablation to demonstrate the effectiveness of our method. 
We compute the average values of the lines parallel to the diagonal in the similarity matrix, and find the line that has the largest average value representing the highest correspondence between the streams.
This is done for the diagonal line, and those adjacent to it in the $[-5,+5]$ frame range. This method benefits from the use of similarity matrix and avoids the adverse effect of padding in the baseline sliding window approach, but does not learn to deal with irregular patterns unlike the trainable methods. We name this method \texttt{Diag-avg} in the subsequent tables.

Our approach is more suited to video streaming in a real world scenario. 
The sliding window approach requires audio features to be available for a sizable window around a visual feature.
Therefore, the AV systems must wait while additional audio stream is transmitted, while keeping previous audio features. 
The proposed approach does not wait the additional input or hold extracted features.


\begin{figure}[ht]
  \centering
  \includegraphics[width=0.60\columnwidth]{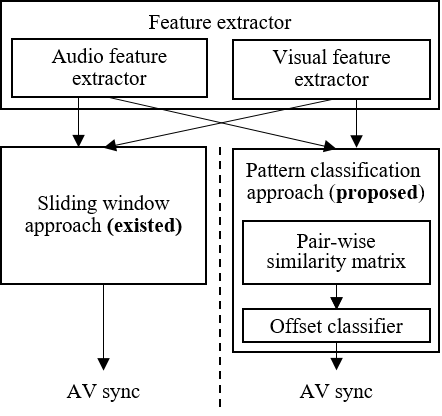}
  \caption{Comparison of existed (left) and proposed (right) approaches. The existed approach determines the AV offset in a heuristic way. 
  While, the proposed approach uses the pattern classifier, which can be trained in two strategies - with the feature extractor or without.}
  \label{fig:speech_production}
\end{figure}

\begin{table}[t]
  \caption{Architecture of \texttt{Sync-cls} network. The second kernel size is same as the input matrix, which is to aggregate the global pattern.}
  \label{tab:word_styles}
  \centering
  \begin{tabular}{ccc}
    \toprule
    \textbf{Layer}  & \textbf{Kernel size}  & \textbf{Filter \#}\\
    \midrule
    Conv1     & (3, 3)  & 256 \\
    Conv2     & (N, N)  & 256  \\
    Conv3     & (1)  & 128\\
    Conv4     & (1)  & 11\\
    \bottomrule
  \end{tabular}
\end{table}


\section{Experiments}

We evaluate the performances of baseline, \texttt{Diag-avg}, \texttt{Sync-cls} and \texttt{Sync-e2e} methods. 
We then analyze the performance improvement given by each of our proposed approaches. 
All the networks in our experiments are implemented using PyTorch~\cite{NEURIPS2019_9015}.

\subsection{Setting}

\subsubsection{Dataset}
{\em Pre-train} sets of Lip Reading Sentences 2 (LRS2)~\cite{chung2017lip} and Lip Reading Sentences 3 (LRS3)~\cite{afouras2018lrs3} are used for training and testing.
LRS2 has 96,318 clips in the pre-train set and consists of videos sourced from the BBC television. LRS3 has 118,516 clips which are sourced from TED and TEDx videos. 
Both sets are used for training, however 7,100 videos from the LRS2 dataset is reserved for testing.

All videos in the dataset are perfectly synchronised, therefore we introduce a random artificial offset of $[-5,+5]$ both during training and test.

\subsubsection{Evaluation protocol}


Our evaluation setting is more strict compared to~\cite{chung2019perfect, chung2020perfect}.
They use additional audio frames in making the window, and see more temporal information using the wider window. 
We do not use the additional input, and determine an offset only using the window given for both audio and video. 
Therefore, whereas their evaluation protocol for a 15-frame input looks at 15 video frames and 45 audio frames, our protocol only ingests 15 video frames and 15 audio frames which is more realistic for short-utterance synchronisation.
\cite{chung2019perfect, chung2020perfect} measure accuracy with a \(\pm1\) video frame tolerance. 
With this setting, a prediction is regarded as correct when it falls within the \(\pm1\) range of the ground truth.
For more precise evaluation, we measure 2 types of accuracies -- with the \(\pm1\) frame tolerance or without.

We repeat the evaluation 10 times and report the average value and standard deviation to mitigate the effect of random offset during evaluation.

\subsection{Result}

\subsubsection{Performance evaluation}

Table 2 shows the accuracy without the one frame tolerance for different numbers of input RGB frames.
Baseline shows the lowest performance in all cases, and the difference is particularly large for shorter clips.  
\texttt{Diag-avg} shows the significant improvement over the baseline.
\texttt{Sync-cls} predicts the time offset better than both non-trainable methods, because of its robustness to the noisy patterns that result from less representative features or non-informative video segments. 
\texttt{Sync-e2e} consistently shows the highest performance across all number of the input frames, and the best accuracy without tolerance is $79.26\%$ with 20 input images. 
The end-to-end method shows superior performance to \texttt{Sync-cls} since the whole network is trained for the task, hence the feature extractor is able to provide optimal features for the classifier to predict the offset. 

Table 3 shows the accuracy with one frame tolerance for all methods. The overall tendency is similar to the case without tolerance. The baseline shows the lowest accuracy, and \texttt{Sync-e2e} achieves the highest performance across all scenarios. 
The best model correctly predicts the offset of $95.18\%$ of videos with one frame tolerance, using only 0.8 second streams. 

We calculate the relative error reduction (RER) between the baseline and the proposed methods. 
Across all experiments, the proposed methods show significant improvement over the baseline, with the highest relative gain of $39.59\%$ for 11 frames in the experiments without tolerance. 
The RER is more significant with tolerance than without, and the biggest error reduction rate is $50.22\%$ for 11 frame input.
Note that the relative performance gain is smaller when the input length is longer -- this is because the baseline method can already predict most of the examples correctly, while many of incorrectly classified segments contain no discriminative information.

\begin{table}[ht]
\caption{Accuracy (\%) \textbf{without} tolerance and RER. For accuracy, we experiment $10$ times and report the mean and the standard deviation. RER in a bracket denotes relative error reduction rate between baseline and the proposed method.}
\label{tab:res}
\resizebox{\columnwidth}{!}{\begin{tabular}{ccccc}
\toprule 
\# RGB frames   & 11                                                                & 13                                                                & 15                                                                & 20                                                                \\ \hline
Baseline  & \begin{tabular}[c]{@{}c@{}}45.17\\ $\pm$0.32\end{tabular}         & \begin{tabular}[c]{@{}c@{}}56.46 \\ $\pm$0.36\end{tabular}        & \begin{tabular}[c]{@{}c@{}}63.66 \\ $\pm$ 0.49\end{tabular}       & \begin{tabular}[c]{@{}c@{}}73.66 \\ $\pm$ 0.33\end{tabular}       \\ \hline
\texttt{Diag-avg}     & \begin{tabular}[c]{@{}c@{}}63.34\\ $\pm$0.39\\ (31.32)\end{tabular} & \begin{tabular}[c]{@{}c@{}}67.46\\ $\pm$0.15\\ (25.26)\end{tabular} & \begin{tabular}[c]{@{}c@{}}71.91\\ $\pm$0.29\\ (22.70)\end{tabular} & \begin{tabular}[c]{@{}c@{}}77.95\\ $\pm$0.27\\ (16.27)\end{tabular} \\ \hline
\texttt{Sync-cls} & \begin{tabular}[c]{@{}c@{}}64.12\\ $\pm$0.49\\ (34.56)\end{tabular} & \begin{tabular}[c]{@{}c@{}}69.43\\ $\pm$0.36\\ (29.79)\end{tabular} & \begin{tabular}[c]{@{}c@{}}73.69\\ $\pm$0.25\\ (27.59)\end{tabular} & \begin{tabular}[c]{@{}c@{}}79.02\\ $\pm$0.24\\ (20.34)\end{tabular} \\ \hline
\texttt{Sync-e2e} & \begin{tabular}[c]{@{}c@{}} \textbf{66.88}\\ $\pm$0.35\\ (39.59)\end{tabular} & \begin{tabular}[c]{@{}c@{}} \textbf{70.76}\\ $\pm$0.44\\ (32.84)\end{tabular} & \begin{tabular}[c]{@{}c@{}} \textbf{74.38}\\ $\pm$0.39\\ (29.50)\end{tabular} & \begin{tabular}[c]{@{}c@{}} \textbf{79.26}\\ $\pm$0.29\\ (21.25)\end{tabular} \\ \bottomrule

\end{tabular}}
\end{table}

\begin{table}[ht]
\caption{Accuracy (\%) \textbf{with} tolerance and RER. For accuracy, we experiment $10$ times and report the mean and the standard deviation. RER in a bracket denotes relative error reduction rate between baseline and the proposed method.}
\label{tab:res}
\resizebox{\columnwidth}{!}{\begin{tabular}{ccccc}
\toprule 
\# RGB frames   & 11                                                                & 13                                                                & 15                                                                & 20                                                                \\ \hline 
Baseline  & \begin{tabular}[c]{@{}c@{}}76.73\\ $\pm$0.34\end{tabular}         & \begin{tabular}[c]{@{}c@{}}83.85 \\ $\pm$0.40\end{tabular}        & \begin{tabular}[c]{@{}c@{}}88.05 \\ $\pm$ 0.26\end{tabular}       & \begin{tabular}[c]{@{}c@{}}93.34 \\ $\pm$ 0.23\end{tabular}       \\ \hline
\texttt{Diag-avg}     & \begin{tabular}[c]{@{}c@{}}85.05\\ $\pm$0.35\\ (35.73)\end{tabular} & \begin{tabular}[c]{@{}c@{}}88.38\\ $\pm$0.25\\ (28.08)\end{tabular} & \begin{tabular}[c]{@{}c@{}}90.88\\ $\pm$0.20\\ (23.71)\end{tabular} & \begin{tabular}[c]{@{}c@{}}94.35\\ $\pm$0.08\\ (15.08)\end{tabular} \\ \hline
\texttt{Sync-cls} & \begin{tabular}[c]{@{}c@{}}86.52\\ $\pm$0.19\\ (42.05)\end{tabular} & \begin{tabular}[c]{@{}c@{}}90.49\\ $\pm$0.24\\ (41.10)\end{tabular} & \begin{tabular}[c]{@{}c@{}}92.10\\ $\pm$0.19\\ (33.92)\end{tabular} & \begin{tabular}[c]{@{}c@{}}\textbf{95.18}\\ $\pm$0.18\\ (27.64)\end{tabular} \\ \hline
\texttt{Sync-e2e} & \begin{tabular}[c]{@{}c@{}}\textbf{88.42}\\ $\pm$0.29\\ (50.22)\end{tabular} & \begin{tabular}[c]{@{}c@{}}\textbf{90.79}\\ $\pm$0.15\\ (43.00)\end{tabular} & \begin{tabular}[c]{@{}c@{}}\textbf{92.60}\\ $\pm$0.19\\ (38.06)\end{tabular} & \begin{tabular}[c]{@{}c@{}}\textbf{95.18}\\ $\pm$0.14\\ (27.51)\end{tabular} \\ \bottomrule

\end{tabular}}
\end{table}

\subsubsection{Analysis}
\begin{figure}[ht]
  \centering
  \includegraphics[width=0.8\linewidth, trim = 5 7 0 0, clip]{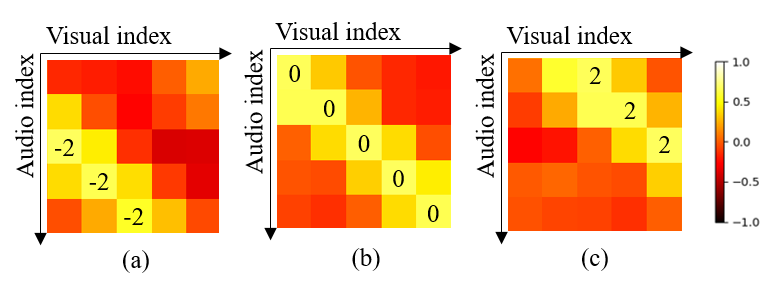}
\caption{Expectation of the relation between pattern on similarity matrix and AV offset. AV offset of (a), (b), and (c) are $-2$, $0$, and $2$ respectively.}
\label{fig:5}
\end{figure}

In this section, we analyze the behavior of the proposed end-to-end learning framework. 
In Figure~\ref{fig:5}, we visualise the expected relationship between the location of linear pattern and the AV offset.
Three different cases are represented, and the AV offset of each sample is denoted on the pattern. 
As shown in Figure~\ref{fig:5}a, if the AV offset is $-2$, then the linear pattern is located two steps below the diagonal. If the AV offset is $0$, the linear pattern is on the diagonal of the matrix (Fig~\ref{fig:5}b), and if the offset is $-2$, then the pattern is two steps above the diagonal (Fig~\ref{fig:5}c). 

\begin{figure}[ht]
  \centering
  \includegraphics[width=0.8\linewidth, trim = 7 7 0 5, clip]{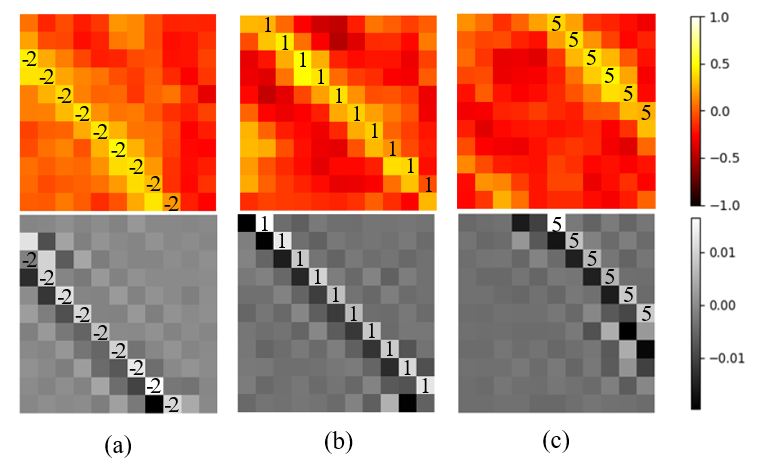}
\caption{Example of similarity matrix (top) and its gradient (bottom). AV offset of (a), (b), (c) are $-2$, $1$, and $5$.}
\label{fig:6}
\end{figure}

We analyze the back-propagated gradients in order to identify important elements or regions for models to classify offset from an input matrix. 
In particular, the gradient values on the similarity matrix are back-propagated from the output of the classification network using true offset. 
A high value of gradient calculated at a particular point means that the change at that point has a significant effect on the network's output. 
This type of analysis is used to identify salient features in image classification neural networks~\cite{Simonyan14a}. 
Figure~\ref{fig:6} shows examples of similarity matrix (top) and its gradient (bottom). The ground truth label of (a), (b), (c) are $-2$, $1$, and $5$ respectively. Examples are generated using \texttt{Sync-e2e} that shows the highest accuracy, and gradients are back-propagated from the node of the output layer to the similarity matrix. 
Most of the gradient are close to $0$, and hence do not affect the offset classification. 
On the other hand, the gradient of the region related to AV offset has the highest value.
The gradient on the line that exactly matches the AV offset have high positive values (represented by white color), indicating that these elements affect the correctly classified output. 
These results demonstrate that the model learns the position of the linear pattern to predict the AV offset, following our expectations indicated earlier. 
In addition, high negative gradients (represented by black color) on the region adjacent to the true offset enable accurate offset prediction.

\begin{figure}[ht]
  \centering
  \includegraphics[width=0.8\linewidth, trim = 5 7 0 5, clip]{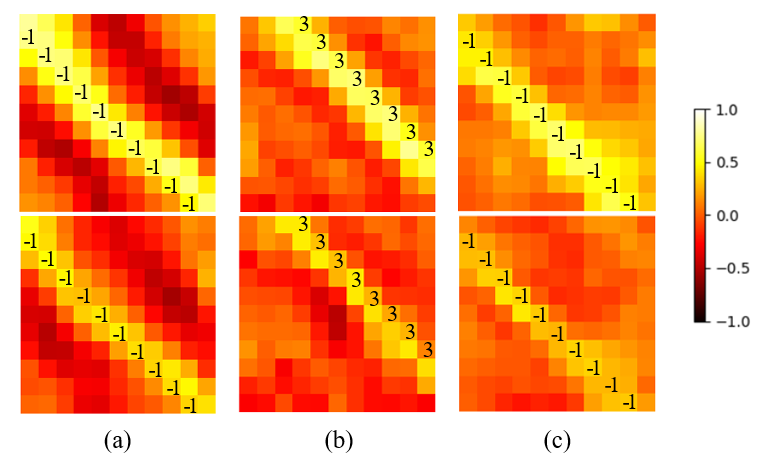}
\caption{Comparison of similarity matrix between before (top) and after (bottom) \texttt{Sync-e2e} learning. 
\texttt{Sync-e2e} learning generates narrow pattern, which results in more accurate prediction.}
\label{fig:7}
\end{figure}

We compare the pattern on the similarity matrix between before and after the end-to-end training, in order to identify the benefits and the performance gains of the end-to-end learning. 
Note that \texttt{Sync-cls} does not affect the similarity matrix.
Examples are visualised in Figure~\ref{fig:7}, and the upper row shows the similarity matrices produced before \texttt{Sync-e2e} learning. The truth label of (a), (c) are $-1$, and (b) is $3$. 
\texttt{Sync-e2e} produces relatively narrow pattern where the two streams are synchronised, which reduces the possibility of incorrectly classifying the pattern into adjacent classes.
We conclude that gradients from the classification loss (cross-entropy loss) flows to the feature extractor, encouraging the network to learn more optimised features for temporal offset prediction. 
The similarity matrix derived from the tuned features has more compatible patterns for offset classification, and as a result \texttt{Sync-e2e} achieves higher accuracy.






\section{Conclusion}


In this paper, we have proposed a new approach to synchronise audio and video in a talking face. 
Previous works extract the audio and visual features using the trained feature extractor, and then determine the time offset between the two streams by using the sliding window approach.
However, there are many challenging cases that are difficult to handle using the heuristic in the previous works.
To this end, we propose a novel approach that trains the network to directly find the time offset between audio and visual streams in an end-to-end manner. 
We first calculate similarities between audio and visual features to construct the similarity matrix, and then regard the matrix as an image and find the pattern to classify the time offset between the two streams. The feature extractor and the classifier are trained end-to-end.
We evaluate the performance of the baseline, \texttt{Diag-avg}, \texttt{Sync-cls}, and \texttt{Sync-e2e}, using LRS2 and LRS3 lip-reading datasets.
Proposed methods outperform the baseline by a large margin, across a range of experiments.
Our best model shows the highest accuracy 95.18\%, only given a short input of 0.8 seconds.

\section{Acknowledgements}
We would like to thank Joon Son Chung, Jaesung Huh, Seongkyu Mun and Jingu Kang at Naver Corporation for their helpful advice.

\bibliographystyle{IEEEbib}
\bibliography{longstrings,mybib,vgg_local,vgg_other}

\end{document}